\documentclass[a4paper]{sbgames}               % final
\usepackage{times}
\usepackage{graphicx}

\usepackage{parskip}

\usepackage[labelfont=bf,textfont=it]{caption}

\usepackage{url}

\title{A Tutor Agent for MOBA Games}

\author{Victor do Nascimento Silva and Luiz Chaimowicz\\ \\Departament of Computer Science, Universidade Federal de Minas Gerais, Brazil}
\contactinfo{\{vnsilva,chaimo\}@dcc.ufmg.br}

\begin{document}

%\teaser{
%  \includegraphics[width=\linewidth]{sample.pdf}
%  \caption{Optional image}
%}

%% The ``\maketitle'' command must be the first command after the
%% ``\begin{document}'' command. It prepares and prints the title block.

\maketitle

%% Abstract section.

\begin{abstract}
Digital games have become a key player in the entertainment industry, attracting millions of new players each year. In spite of that, novice players may have a hard time when playing certain types of games, such as  MOBAs and MMORPGs, due to their steep learning curves and not so friendly online communities. In this paper, we present an approach to help novice players in MOBA games overcome these problems. An artificial intelligence agent plays alongside the player analysing his/her performance and giving tips about the game. Experiments performed with the game {\em League of Legends} show the potential of this approach.
\end{abstract}

%% The "\keywordlist" command prints out the keywords.
\keywordlist
\contactlist

\section{Introduction}

Multiplayer Online Battle Arena (MOBA) games have become very popular in recent years, reaching millions of users everyday. On a first sight, MOBA games present a simple game concept: the player controls a hero that should use its abilities to fight enemy heroes and creeps, destroy structures and conquer an enemy base. 
Despite this simple concept, this game genre have a complex gameplay and requires a lot of knowledge that may not be so familiar to novice players\cite{yang2014identifying}. They should understand how to walk, cast spells, last hit creeps, harass enemies, fight neutral creeps, attack, perform itemization, among other things. Resource management must be done, like using mana, energy and gold, and players should understand how to build his/her hero and plan strategies based on these features. 

% "There is no evaluation of learning": isso é importante no contexto do artigo? nós atacamos esse ponto? Achei mehor tirar a frase

Games often present a tutorial where the main gameplay is presented, aiming to teach newcomers how to play. Requiring the player to complete tutorial matches or playing against bots is also a very common practice, trying to assure that the player understands the concepts presented. However, in MOBA games, there are some features or roles that require the cooperation of more than one player. To group players of similar skills in a team, these games implement a matchmaking system \cite{mylak2014developing}, but, in a cooperative scenario, this approach may not perform well. The performance can be particularly poor in initial matches, as the combination of players with low skill or no knowledge of some features in the same team will not help them in learning the game. 

%there is no evaluation of learning, and

In this paper we propose an approach to help novice players to learn the basics of MOBA games, improving their entertainment. The idea is to develop an Artificial Intelligence Agent that will play alongside a human team acting as a tutor. This agent will provide tips and support players, aiming at improving their experience and gameplay while also benefiting the game by reducing the number of dropouts. This tutor will replace one of the human players and will go unnoticed by others, which will consider it a regular player. Despite of the fact that the tutor supports all team, it is focused in supporting a specific player, that it identifies as partner.

The rest of this paper is organized as follows: Section 2 presents some definitions related to MOBA games and their player community while in Section 3 we discuss some related works. Section 4 describes the mechanisms used for the tutor development and Section 5 presents the  experiments and obtained results. Finally, Section 6 brings the conclusion and directions for future work.

\section{MOBA Games}

Before describing our approach, it is important to present the gameplay, characteristics and features of MOBA games. We also discuss the profile of the gamer that generally plays this game genre, presenting its behavior in community.

\subsection{General Description}

Multiplayer Online Battle Arena is a genre originated from Real Time Strategy (RTS) games, in which players control an unit (Hero) in a map, taking part in a battle to destroy an enemy base, like a capture the flag game. As the popularity of RTS games increased in the last decade, players began to develop custom maps and features to games, called MODs, using tools distributed with the game. Thus, a custom map from \emph{Starcraft} was the start point of MOBA games, giving rise to \emph{Aeon of Strife}. The popularity of this genre grew and users from similar games began to develop similar maps. \emph{Defense of the Ancients} (DotA) was one of these maps, originated from \emph{Warcraft III}. DotA was so successful that its name was used to describe the genre for a long time, the DotA like games. The term MOBA just came up in 2009, when Riot Games used it to describe its debuting title \emph{League of Legends} (LoL) \cite{ferrari2013generative}. Therefore, MOBA became very popular and originated several other titles like \emph{Dota 2}, \emph{Heroes of the Storm}, \emph{Heroes of Newerth} and \emph{Strife}. Data from Riot Games shows that 67 millions of users plays matches every month, with 27 millions of players daily and 7.5 millions of player simultaneously \cite{riotLoLdata}.

MOBA gameplay generally consist of matches where ten players compete, five in each team. At the beginning of a match, each player must choose a hero, which he/she will be playing during the entire match. The hero is a powerful unit with an unique spell set that can be used to build complex game strategies and goals. Controlling that hero, the player should join his teammates in a competition to destroy an enemy base, which is heavily guarded by turrets and structures \cite{whatsMoba}. A MOBA map is generally composed of three lanes: bottom, mid and top, where AI driven creeps spawns. By defeating these creeps, neutral creeps or enemy heroes the player can gather gold and experience. While gold can be used to buy items in order to improve hero status, experience makes the hero gain level, getting stronger and learning new abilities.

\subsection{Community}

Real Time Strategy players are widely known by their expertise in handling lots of information simultaneously and for also being very ``toxic'', raging at novice players \cite{kwak2015exploring}. In {\em Warcraft III} it is very common to kick players from matches when they are identified as novice. Since {\em DotA} evolved from {\em Warcraft III} it is also possible to kick players from games. When {\em DotA} became popular in online environments, the common sandbox were hold in a P2P architecture, where one player is the host and other players are able to connect to his match. As result, most players were able to join matches where there are players that already know the game environment and mechanics very well. Instead of being friendly and patient to newcomers, experienced players usually  bully them \cite{kwak2015exploring},\cite{blackburn2014stfu}.

Raging characteristics from RTS players seems to be connected to modern MOBA games as most of them migrated from RTS to MOBA. Valve approach to fight toxic behaviour in {\em DotA 2} uses a player report system. Riot also noticed toxic behavior amongst LoL players, having initially created a system where players could judge cases based on reports \cite{lin2013}. However they recently updated their system, applying Machine Learning Techniques to identify toxic behavior\cite{lin2015}.

As presented, MOBA community its not very friendly \cite{blackburn2014stfu}, and that could imply in players feeling frustrated more often. Some online games have similar behavior amongst their players, and try to solve the problem by encouraging experienced players to help newcomers. In these games, to avoid the experience level gap, developers or even the community implements an "adopt a newbie" program, where an experienced player is encouraged to help a newcomer finishing quests, gathering items and getting levels. When the newcomer becomes experienced both, the godfather and the newcomer, can be gifted in game with special items, gold or titles in the community.

Csikszentmihalyi presents a model for identifying if players are in flow \cite{csikszentmihalyi2014flow}. Being in flow state means that players must not feel frustrated or bored, following in a way that the entertainment is reached, having a good game environment and community can help players to be in flow state. However MOBA games involves much more than just the difficulty presented by game mechanics. Some human factors, like toxic behavior, can drive players to frustration. Playing in team requires cooperation and patience, and that is what the matchmaking system from MOBA games tries to obtain \cite{mylak2014developing}. Matching players of similar skills could provide a friendly environment, where players could grow together, in skills and knowledge, but that hardly happens if there is no one that can teach. It is not difficult to find players appealing to video tutorials or pro player guides trying to learn game features. Also it is hard to reach flow state if players need to leave game environment to watch or read tips. It is even harder to reach flow when there is another player flaming at you by your mistakes or inexperience. 
% ***** Inverter seções 2 e 3?
% ***** Talvez mudar essa frase para os Trabalhos Relacionados, criando um parágrafo com o que o nosso trabalho se %diferencia dos outros e as contribuições.

\section{Related Work}

Tutorials are the main tool to interactively teach game features to players in modern games. Normally, these teaching approaches are simpler to apply since the game provides an environment for practicing what the player has learned \cite{vn2012investigaccao}. Although most game tutorials are successful to show simple game features, it is not hard to find players that claim to have learned less than necessary or did not understand the features at all. This can be justified by the complex gameplay and the broad range of mechanics found in modern games.

%Tutorials is such a new feature in games, in 90s it wasn't common to find games that teaches how to play interactively. In games like Sonic or Mario, players had to read a manual or just figure out how to perform actions or combos. Modern games have a lot of information and a large amount of controls, despite to few buttons from old joysticks found back in 90s. The information amount is especially large when we analyse games like RTS, MMO, RPG or MOBA, the last one is the main target of this paper. 

MOBAs also have joined the tutorial phenomena, as they are generally harder to learn than simple games. In {\em Dota 2}, for example, there is a tutorial in which the player has a tutor hero that guides him/her through the main game features and mechanics such as turrets, creep farming and denying. {\em Strife} and {\em Heroes of Newerth} both offer similar environments of learning main gameplay. By defeating a boss, that is a champion, the player completes the dungeon like tutorials. {\em League of Legends} offers the player a tutorial that shows the game features in an single lane dungeon. All tutorials cited above require and/or invite the player to join a set of matches against AI in order to improve his/her skills and be able to play against other players.
There is also some academic research on how to use AI mechanisms to help players to learn complex games. For example,  in Cunha et. al. \cite{cunha2015rtsmate}, authors implemented an advisory system that gives hints based on the current status of the game and the performance of the player. They tested their approach in {\em Wargus}, a RTS game with a complex scenario. 
%***** Seria interessante citar outros trabalhos de AI Tutorials se houver
On a more broad sense, a general survey on the use of real time AI based teammates in several games genres can be found in \cite{mcgee2010real}. AI teammates can be used with different purposes, including to helping the players to learn a game, as we do in this paper.

% esse último parágrafo poderia estar na introdução / moticação, mas podemos deixar aqui memso, ou eventualmente juntar com o próximo que deve entrar...
The basics of game features must be presented to players in every MOBA game, and some of these games only show the mechanics in a very high level. Thereby, most players claim that the tutorials are not sufficient to teach the basics of the game. Moreover, it normally takes time to initially start playing a tutorial, then playing with bots and finally Player versus Player (PvP). So, players generally become impatient and go straight to the PvP game. And it is not difficult to find player that gets frustrated by meeting experienced players playing in low level accounts, called ``smurfs''. So most players tend to search for partners who can teach them the game properly, others does not even keep trying, just stop playing.

%***** Poratanto o principal diferencial aqui é....

Instead of just giving a set of instructions to the player, our approach introduces an AI agent that plays alongside him/her. The agent aims to teach the player the general game features giving instructions while he/she plays the game. At first that may just seems like a regular game tutorial but, in our approach, the agent analyses the player skills and teaches him/her specific features verifying the player performance. Furthermore, the agent is capable of helping the player to improve his/her skills in game using a game character. This approach reduces the player frustration by ensuring that he/she is playing with a reliable, non-toxic partner.

\section{Implementing a Tutor Bot for MOBA Games}

The first step in our implementation was to find a suitable MOBA platform. Most of them are commercial games and do not provide tools that allow developers to implement novel game features or mods. After doing a broad research we found that the best option was to implement our system in a game that is popular and stable, so we chose {\em League of Legends} (LoL) as our main platform. We then chose to use third party tools to develop the tutor system. Thereby we develop our approach using Bot of Legends (BoL) a third party tool that injects scripts written in Lua in the game. BoL is maintained by an open developers community, and only has access to the information the players have. (cheating isn't allowed). 

% o que é um Metagame?
%Metagame é o uso de informações de fora do jogo para a definição de uma estratégia dentro do jogo, porém no nosso caso o termo estratégia se encaixa bem.
In MOBA there are different roles the heroes can assume, each one aiming to fulfill some needs of the team, given a strategy. We identified that the best role to help players improve their skills is {\em support}. This role is characterized by the presence of utility spells, such as cure, shields or disables, that can improve the ally hero survival potential. The support is normally responsible for helping a \emph{carry} (champion that does a great amount of damage) in the bottom lane, in the current strategy of LoL. After the lane phase, it is responsible by helping all players in team fights and lane pushing. We chose {\em Soraka}, a LoL support champion that easily fills the needs of a novice player, providing support and healing while he/she learns how to play the game. {\em Soraka} is focused in high crowd control, having spells that slow, root and silence enemies in area-of-effect. She also has spells that can cure close allies and, lastly, her ultimate spell can cure the entire team. {\em Soraka} passive spell allows her to run faster towards injured allies, making her helpful not just for one player, but for the entire team.

We then developed a script that selects the bottom lane player, as supporters normally go in the bottom lane, and follows the player from an adjustable distance, avoiding collisions with him/her. This system uses a two layered architecture: the first layer is responsible for driving the agent to follow the player while the second layer controls the use of skills to support the player. The support system only helps the player, supporting him during the entire game. However the support system does not give tips to player about mechanics and techniques. The tutor system tracks the player mechanics and gives tips to improve it. We should stress that our goal is to develop an agent capable of helping players to develop their skills, thereby is not our current goal to develop an agent that is Turing acceptable. We developed a message system that is integrated with LoL smart ping system. It emits messages and smart signals warning the player when he/she has low health or bad positioning. A rule based system is used to evaluate the data collected, getting the tips from a lookup table. The table tip was developed based on the video guides from Riot Games and also on texts from LoL champion pages. For positioning, we give tips to the player when he/she is going to be focused by towers, enemies or minions. These specific tips are given to all players in the team. 

\begin{figure*}
\centering
\includegraphics[width=\textwidth]{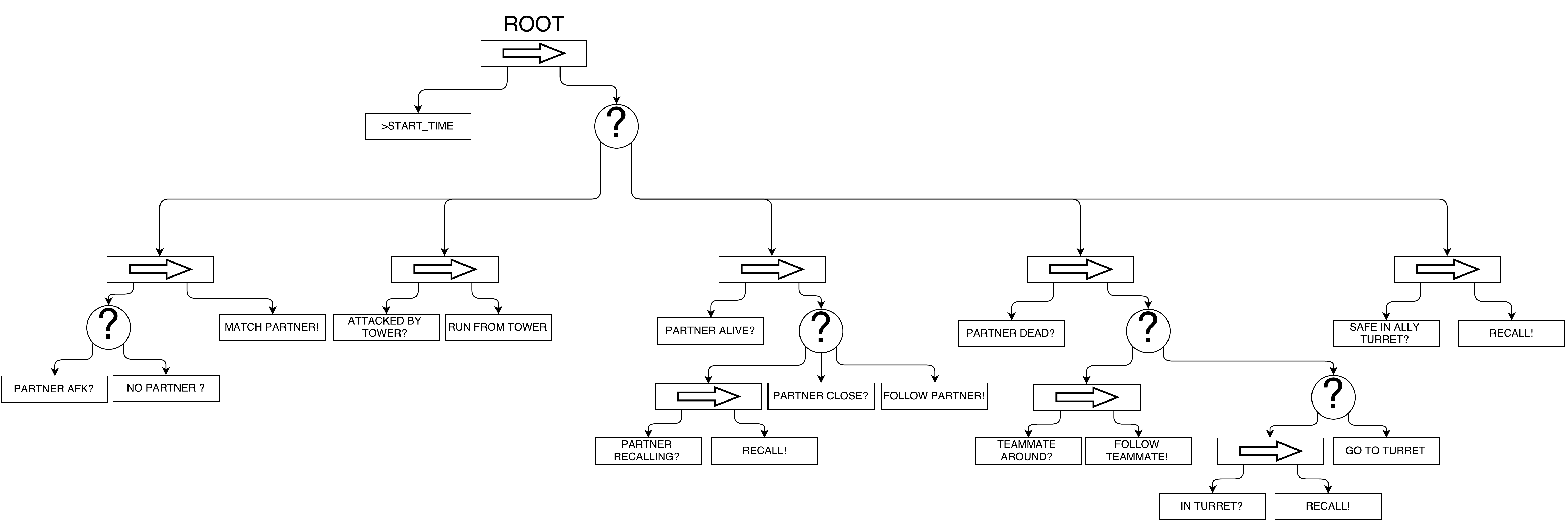}
\caption{Behavior Tree implemented by the agent. The Sequencers are represented by arrow rectangles, Selectors by question circles. Actions and questions are represented by texts inside rectangles.}
\label{fig:BTSoraka}
\end{figure*}

%Added
For developing the agent we extracted domain knowledge from expert players, in-game tutorials, videos and streaming from expert players. We then divided this information in tactical and positioning, attributing them to movement and mechanism layer respectively. The movement system was developed using the knowledge modelled in a Behavior Tree, than can be seen in Figure \ref{fig:BTSoraka}, used to decision making. Real time reasoning was used to cause stimuli in the Behavior Tree and perform the decision making strategy.

\section{Experiments}

%To test our approach, we performed some experiments introducing our tutor as a player in a match. These experiments are presented below.

\subsection{Experiment 1: Playing alongside human players}

We run it in a low level account, playing in cooperative versus AI mode, cooperating with 4 human players. League of Legends provides players a match history, where you can retrieve matches from accounts by visiting their profile. It is also possible to record information from the after match screen, where score information is shown for each player who took part in the match. A brand new, level 1 account was created assuring that LoL matchmaking would select players with zero or low game knowledge. 

For data collection, we tracked the player that partnered with the tutor. We compared the results of three matches played prior and three matches played right after the match played with the tutor. Also, we ran 3 matches using only the supporting system and 3 matches with supporting and tip system enabled. All matches were played in Summoners Rift, Coop vs. AI, Introductory solo queue, with matchmaking selected team. The data has been collected using LoL patch 5.7 in 2015.

As a performance metric, we chose the Kill/Death/Assists (KDA) factor. It consists of analysing the number of enemies killed, the number of times the player helped his teammates to kill an enemy and how many times he/she died. This factor is broadly used to analyse the gameplay and performance of players, both in amateur, competitive or e-Sports scenario.
%We evaluate our approach aiming to improve the average KDA factor of partners, using it as measure if gameplay improvement. 
The KDA factor function is presented in equation \ref{eq:KDAfactor}.

\begin{equation}
    f(KDA) = 
    \left\{\begin{array}{l l}
        \frac{K + A}{D}, & \quad D > 0 \\
        K + A, & \quad \mathrm{else}
    \end{array} \right.
    \label{eq:KDAfactor}
\end{equation}

\begin{figure}[t]
\centering
\includegraphics[width=9.20cm]{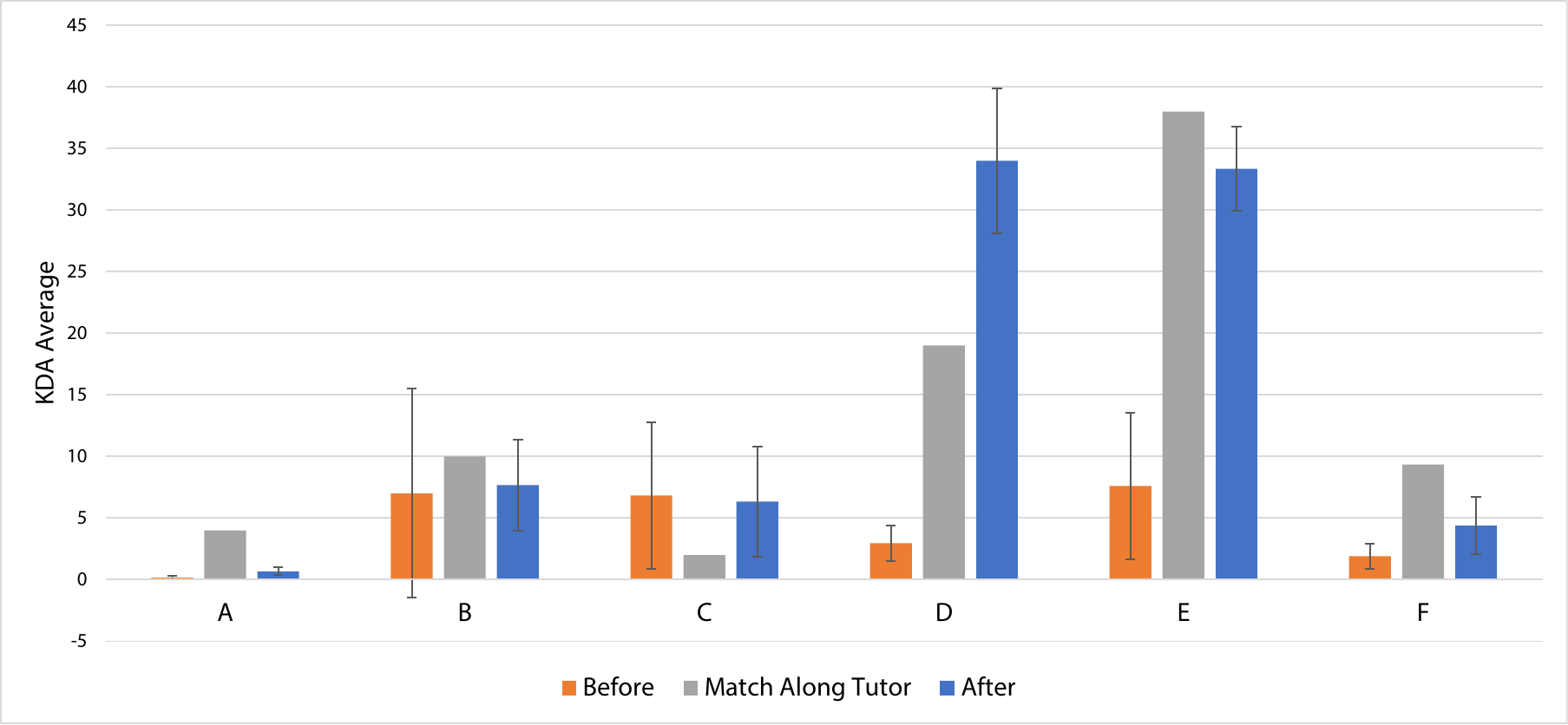}
\caption{KDA performance of six players (A-F) before and after playing with the tutor.}
\label{fig:KDAAnalysis}
\end{figure}

We performed experiments with 6 different players (A-F). Players A, B and C played matches with the support system only, while D, E and F played with support and tip system enabled.
For each player, we computed the average and standard deviation of the KDA factor from 3 matches before and after the match played with the tutor. The results are displayed in Figure~\ref{fig:KDAAnalysis}. 

% Análise Tutor x Tutor+Tip / Melhora do KDA durante e depois da partida
When comparing the matches using only the support system (A,B,C) with support plus tip system (D,E,F) we can observe that players which used the tip system had a larger performance improvement in matches played right after playing with the tutor. This large improvements are not observed when only the support system is used (A,B,C). 
In the match played with the tutor (gray bar), all players, except C, showed better than average performance, supporting our hypothesis that the companion of a more knowledgeable player improves performance. We can also observe that players D, E, F showed much better performance when playing along our tutor, showing that the tip and support system together is helpful to new players. Overall, we observe that all players, except by C, showed performance improvements after playing along the tutor , showing that the tutor may help them to improve their gameplay. Player C presents a slightly decrease in performance, and that can be caused by various factors, such as bad connection, player experience and others.

%Standard deviation analysis
We also observed that some players presented a high standard deviation prior to play a match along the tutor, meaning their performance was unstable. As all players were under level 10, that can be considered normal, as these players are learning how to play. Results shown that players B, C and E reduced the standard deviation, meaning they became more stable in their gameplay. On the other hand, we observed that players A, D and F showed higher standard deviations after playing along the tutor. Standard deviation, in our data collection, can be influenced by many factors like the performance increase, player experience, connection problems, and others, that means it should not be taken as a conclusive evaluation factor in this study.

Finally, League of Legends provides a chat room in game and after it. It also provides an honor system, in which partners can honor teammates for being helpful, friendly or showing teamwork. Moreover, LoL lets users add others as friends, so they can invite others to play matches together. After the three matches with tip system our tutor got three honors, one for being helpful and two for teamwork. It also got one friend request and some chat compliments. %We does not consider chat compliments, as it can't be retrieved.

%Luiz, eu não apaguei a seção de experimentos anteriores apenas adicionei estas por não sei se iremos reaproveitá-la. 
\subsection{Experiment 2: Survey and play with controlled players}

For evaluation of our approach we performed some experiments with players. We invited random players to join our agent in two matches. In the first one the player would play alongside the agent without the tip system. In the second match, the player would play with the tips on. For information collection we applied two surveys\footnote{Survey 1: http://bit.ly/1HJtBDg\\Survey 2: http://bit.ly/1g5HvqA}. First survey was applied before the matches was actually played, aiming to collect information about our testers. We performed early tests with two users for evaluation and fixing possible problems, the results were discarded.

We have a set of six players, all male with ages ranged from 18 to 24 years old. In the first survey we collected data from their experience with MOBA and League of Legends. The experience of these players could reveal how well they could evaluate our system. Also, we have asked players about their experience with MOBAs others than LoL, evaluating the overall expertise of these players. 

All players were familiar with MOBA, three of them were advanced, two intermediate and one beginner. We notice that the majority or players are advanced, and we expect these players could provide us valuable feedback about the tutor. In addition, all of the players surveyed were main players of LoL and had experience with at least one MOBA game out of LOL.

Further we asked players about their experience in LoL with the initial game tutorial. We asked players to attribute a score to the tutorial ranged from 0 to 10, where 0 means very unsatisfied and 10 very satisfied. Our results found that most players consider the tutorial of median to poor value. We had two players that did not do the tutorial, two that attributed a score of four and two that gave a score of five to the LoL tutorial. In addition we asked if the player considers that the tutorial is sufficient for teaching a new player the basic concepts of the game obtaining. We also asked questions about the hero, the game and player experience, however we will not include all details in this work due to paper size.

After the matches played alongside our agent, we asked the player to answer a survey about his experience with the agent alone and the agent plus the tip system. The survey revealed that none of the players has played with a system like the one we present in this work. We asked players question about the performance of the agent in the match. We also asked players how they thought that the agent would help novice players and how good the agent was, shown in Table \ref{table:helper}. Further, we asked about the tip system and the tips frequency, Table \ref{table:frequency}. Lastly, we asked players about the actions that the agent performed during the match, classifying him as robotic or human. Four of these players considered the agent more robotic than human, one considered it very robotic and one considered it nor robotic nor human.

\begin{table}[t]
\centering
\caption{Score of the agent as helper for new players}
\label{table:helper}
\begin{tabular}{lll}
\hline
Score & Number of answers & Percentage \\ \hline
10    & 1                 & 16.6\%     \\
8     & 3                 & 50\%       \\
7     & 2                 & 33.3\%     \\ \hline
\end{tabular}
\end{table}

\begin{table}[t]
\centering
\caption{Answers to: "How do you evaluate the frequency of tips?"}
\label{table:frequency}
\begin{tabular}{lll}
\hline
Answer               & Number of answers & Percentage \\ \hline
Good frequency       & 4                 & 66.7\%     \\
Little low frequency & 1                 & 16.6\%     \\
Low frequency        & 1                 & 16.6\%     \\ \hline
\end{tabular}
\end{table}

\section{Conclusions and Future Work}

In this paper we addressed the problem of helping novice players with the features and mechanics of MOBA games aiming to improve their entertainment. For this, we develop an Artificial Intelligence driven agent that plays along players guiding and tipping them. We ran the agent in League of Legends live servers within a brand new account aiming to find novice players joining the game. This approach showed that players improve their performance when playing with friendly partners that are willing to teach them the game features. In addition, we ran our agent alongside selected players, with set size of six. These players answered two surveys related to their previous experiences related to MOBA and experience with the tutor. They were also invited to play two matches along the agent. Results show promising results towards helping new players, as these players evaluate that the agent performed well as support. They also reported that the frequency of the tips are well balanced in most cases.

For future work, we intend to perform more experiments, both qualitative and quantitative. Also, a deeper analysis can be done by tracking players' progression using the tutor studying the period needed to learn game features and agent lifetime. Another analysis that can be performed is the gameplay improvement of the entire team, measuring tutor efficiency as a multiple supporter. Finally, we want to improve this tutor to teach advanced features, training experienced players. This would help players that already understand the game to improve their skills having a personal coach agent.

\bibliographystyle{sbgames}
\bibliography{template}
\end{document}